\relax

\documentclass[twocolumn]{article}
\usepackage{fullpage}
\usepackage{times}  
\usepackage{helvet}  
\usepackage{courier}  
\usepackage{url}  
\usepackage{graphicx}  
\usepackage{enumerate}

\usepackage[utf8]{inputenc} 
\usepackage[T1]{fontenc}    
\usepackage{url}            
\usepackage{booktabs}       
\usepackage{amsfonts}       
\usepackage{nicefrac}       
\usepackage{microtype}      
\usepackage{color}

\usepackage{graphicx} 
\usepackage{subfigure}

\usepackage{algorithm}
\usepackage{algorithmic}
\usepackage{amssymb}
\usepackage{amsthm}
\usepackage{amsmath,epsfig}
\usepackage{epstopdf}

\usepackage{enumitem}
\newif\ifsubmit

\submitfalse

\ifsubmit
\newcommand{\bo}[1]{}
\else
\newcommand{\bo}[1]{\textcolor{blue}{Bo: #1}}
\fi

\begin{document}
\title{Projection Based Weight Normalization for Deep Neural Networks}

\author{
Lei Huang, Xianglong Liu, Bo Lang\\
Beihang University\\
\texttt{\small\{huanglei,xlliu,langbo\}@nlsde.buaa.edu.cn} \\
\and
Bo Li \\
UC Berkeley\\
\texttt{\small crystalboli@berkeley.edu}
}
\date{}
\maketitle
\begin{abstract}
Optimizing deep neural networks (DNNs) often suffers from  the ill-conditioned problem. We observe that the scaling-based weight space symmetry property in  rectified nonlinear network will cause this negative effect. Therefore, we propose to constrain the incoming weights of each neuron to be unit-norm, which is formulated as an optimization problem over Oblique manifold. A simple yet efficient method referred to as projection based weight normalization (PBWN) is also developed to solve this problem. PBWN executes standard gradient updates, followed by projecting the updated weight back to Oblique manifold.
This proposed method has the property of regularization and collaborates well with the commonly used batch normalization technique.
We conduct comprehensive experiments on several widely-used image datasets including CIFAR-10, CIFAR-100, SVHN and ImageNet for supervised learning over the state-of-the-art convolutional neural networks, such as  Inception, VGG  and residual networks. The results show that our method is able to improve the performance of DNNs with different architectures consistently. We also apply our method to Ladder network for  semi-supervised learning on permutation invariant MNIST dataset, and our method outperforms the state-of-the-art methods: we obtain test errors as $2.52\%$, $1.06\%$, and $0.91\%$ with only 20, 50, and 100 labeled samples, respectively.
\end{abstract}

\section{Introduction}
\label{sec_intro}

Deep neural networks have achieved great success across a broad range of domains, such as computer vision, speech processing and natural language processing~\cite{2015_CVPR_He,2014_ICASSP_Wiesler,2017_AAAI_Tu}. While their deep and complex structure provides them powerful representation capacity and appealing advantages in learning feature hierarchies, it also makes the learning difficult. In the literatures, various heuristics and optimization algorithms have been studied, in order to improve the efficiency of the training, including weight initialization~\cite{1998_NN_Yann,2010_AISTATS_Glorot,2015_ICCV_He}, normalization of internal activation~\cite{2015_ICML_Ioffe}, and sophistic optimization methods~\cite{2015_ICML_Grosse,2017_Corr_Yu}. Despite the progress, training deep neural networks and ensuring satisfactory performance is still considerably an open problem, due to its non-convexity nature and the ill-conditioned problems.

Deep neural networks (DNNs) have a large number of local minima, due to the fact that they usually suffer model identifiability problem. A model is called to be identifiable if a sufficiently large training set can rule out all but one setting of the model's parameters~\cite{Goodfellow-et-al-2016}. Neural networks are often not identifiable because we can obtain equivalent models by swapping their weights with each other, which is called \emph{weight space symmetry} \cite{1993_NC_Chen}. In addition, for the commonly used rectified nonlinear \cite{2010_ICML_Nair} or maxout network \cite{Goodfellow_CoRR_2013}, we can also construct equivalent models by scaling the incoming weight of a neuron by a factor of $\alpha$ while scaling its outgoing weight by $1/ \alpha$. We refer to this as \emph{scaling-based weight space symmetry} \cite{2015_NIPS_Neyshabur}. These issues imply that there can be an extremely large or even uncountably infinite amount of local minima for a neural network. Although it still remains an open question whether the difficulty of optimizing neural networks originates from local minima, we observe that the \emph{scaling-based weight space symmetry} can cause the  Hessian matrix ill-conditioned, which is deemed to the most prominent challenge in optimization~\cite{2010_AISTATS_Glorot,2016_CoRR_Salimans}.

To alleviate the negative effect of \emph{scaling-based weight space symmetry}, we propose to constrain the incoming weights of each neuron to be unit-norm. This simple strategy can ensure that the weight matrix in each layer has almost the same magnitude. Besides, it can keep the norm of back-propagation information during linear transformations. Training neural networks with such constraints can be formulated as an optimization problem over  Oblique manifold~\cite{2006_ICASSP_Absil}. To address this optimization problem, we propose a projection based weight normalization method to improve both performance and efficiency. Our method executes standard gradient updates, followed by projecting the updated weight back to Oblique manifold. We point out that the proposed method has the property of regularization as weight decay \cite{1992_WD_Krogh}, and can be viewed as a regularization term with adaptive regularization factors. We further show that our method implicitly adjusts the learning rate and ensures the unit-norm characteristic for incoming weight of each neuron, under the condition that batch normalization \cite{2015_ICML_Ioffe} is employed in the networks.

We conduct comprehensive experiments on several widely-used image datasets including CIFAR-10, CIFAR-100 \cite{2009_TR_Alex}, SVHN \cite{2011_NIPS_Netzer} and ImageNet \cite{2009_ImageNet} for supervised learning over the state-of-the-art Convolutional Neural Networks (CNNs), such as  Inception \cite{2014_CoRR_Szegedy}, VGG \cite{2014_CoRR_Simonyan} and residual network \cite{2015_CVPR_He,2016_CoRR_Zagoruyko}. The experimental results show that our method can improve the performance of deep neural networks with different architectures without revising any experimental setups. We also consider semi-supervised learning for permutation invariant MNIST dataset by applying our method to Ladder network \cite{2015_NIPS_Rasmus}.
Our method outperforms the state-of-the-art results in this task: we achieve test errors as $2.52\%$, $1.06\%$, and $0.91\%$ with only 20, 50, and 100 labeled training samples, respectively. Code to
reproduce our experimental results is available on: \textcolor[rgb]{0.00,0.50,1.00}{https://github.com/huangleiBuaa/NormProjection}.
Our contributions are as below.
\begin{enumerate}
\item
We propose to optimize neural networks over Oblique manifold, which can alleviate the ill-conditioned problem caused by   \emph{scaling-based weight space symmetry}.
\item
We propose projection based weight normalization method (PBWN), which serves as a simple, yet effective and efficient solution to optimization over Oblique manifold in DNNs. We further analyze that PBWN has the property of regularization as weight decay, and also collaborates well with commonly used batch normalization technique.
\item
We apply PBWN to the state-of-the-art CNNs over large scale datasets, and improve the performance of networks with different architectures without revising any experimental setups. Besides, the additional computation cost introduced by  PBWN is negligible.
\end{enumerate}

%
%
%
%
%
%

\section{Optimization over Oblique Manifold in DNNs}

Consider a learning problem with training data $\mathbb{D}=\{(\mathbf{x}_i, \mathbf{y}_i)\}_{i=1}^{M}$ using a feed-forward neural network $f(\mathbf{x})$ with $L$-layers, where $\mathbf{x}$ refers to the input and $\mathbf{y}$ the corresponding target. The network is parameterized by a set of weights $\mathbb{W}=\{ \mathbf{W}_{l}, 1\leq l \leq {L} \}$ and biases $\mathcal{B}=\{ \mathbf{b}_{l}, 1 \leq l \leq {L} \}$, in which each layer is composed of a linear transformation and an element-wise nonlinearity: $\mathbf{h}_l=\varphi(\mathbf{W}_{l} \mathbf{h}_{l-1}+ \mathbf{b}_l) $. In this paper, we mainly focus on rectifier activation function that has a property of $\varphi(\alpha x)=\alpha \varphi(x)$, and drop the biases $\mathcal{B}$ for simplifying discussion and description.

Given a loss function $\mathcal{L}(\mathbf{y}, f(\mathbf{x}; \mathbb{W}))$ that measures the mismatch between the desired output $\mathbf{y}$ and the predicted output $f(\mathbf{x}; \mathbb{W})$, we can train a neural network $f$ by minimizing the empirical loss as follows:
%
 \begin{eqnarray}
\label{eqn:optimization_normal}
	 \min_{\mathbb{W}} ~~\mathbb{E}_{(\mathbf{x},\mathbf{y})\in \mathbb{D}} [\mathcal{L}(\mathbf{y}, f(\mathbf{x}; \mathbb{W}))].
\end{eqnarray}

In the above formulation, gradient information dominates how to tuning the network parameters. The weight updating rule of each layer for one iteration is usually designed based on Stochastic Gradient Descent (SGD):
\begin{eqnarray}
\label{eqn:update_normal}
	 \mathbf{W}^{*}_{l}=\mathbf{W}_{l} - \eta \frac{\partial \mathcal{L} }{\partial \mathbf{W}_{l}},
\end{eqnarray}
where $\eta$ is the learning rate and the gradient of the loss function with respect to the parameters $\frac{\partial \mathcal{L} }{\partial \mathbf{W}_{l}}$ is approximated by the mini-batch $\mathbf{x}_{1\ldots m}$ of size $m$ by computing $\frac{\partial \mathcal{L} }{\partial \mathbf{W}_{l}}= \frac{1}{m} \Sigma_{i=1}^{m} \frac{\partial \mathcal{L}(\mathbf{y}_i, f(\mathbf{x}_i; \mathbb{W}))}{\partial \mathbf{W}_{l}}$.

\subsection{Scaling-Based Weight Space Symmetry}
In this part, we will  show why the scaling-based weight space symmetry can cause the Hessian matrix ill-conditioned, and this behaviour makes training deep neural network more challenging.



We consider a very simple two-layer linear model with only one neuron per layer, and abuse the rectified nonlinear layer for simplifying discussion without loss of generalization. Let $y=w_2 h_1$ and $h_1=w_1 x$ for the two layers, and define the loss function $\mathcal{L}(y)$. We further assume $w_1$ and $w_2$ are in the same magnitude. Based on the \emph{scaling-based weight space symmetry}, we consider another two-layer linear model parameterized by $ \hat{w}_1= \alpha w_1$ and $\hat{w}_2= \frac{1}{\alpha} w_2$ where $\alpha>1$. Under this parameterization, we can still have the same model output as $\hat{y}=y$ for the same input $x$.

\begin{figure*}[t]
\centering
\hspace{-0.02\linewidth}
 \subfigure[normal parameterizations]{
  \includegraphics[width=0.36\linewidth]{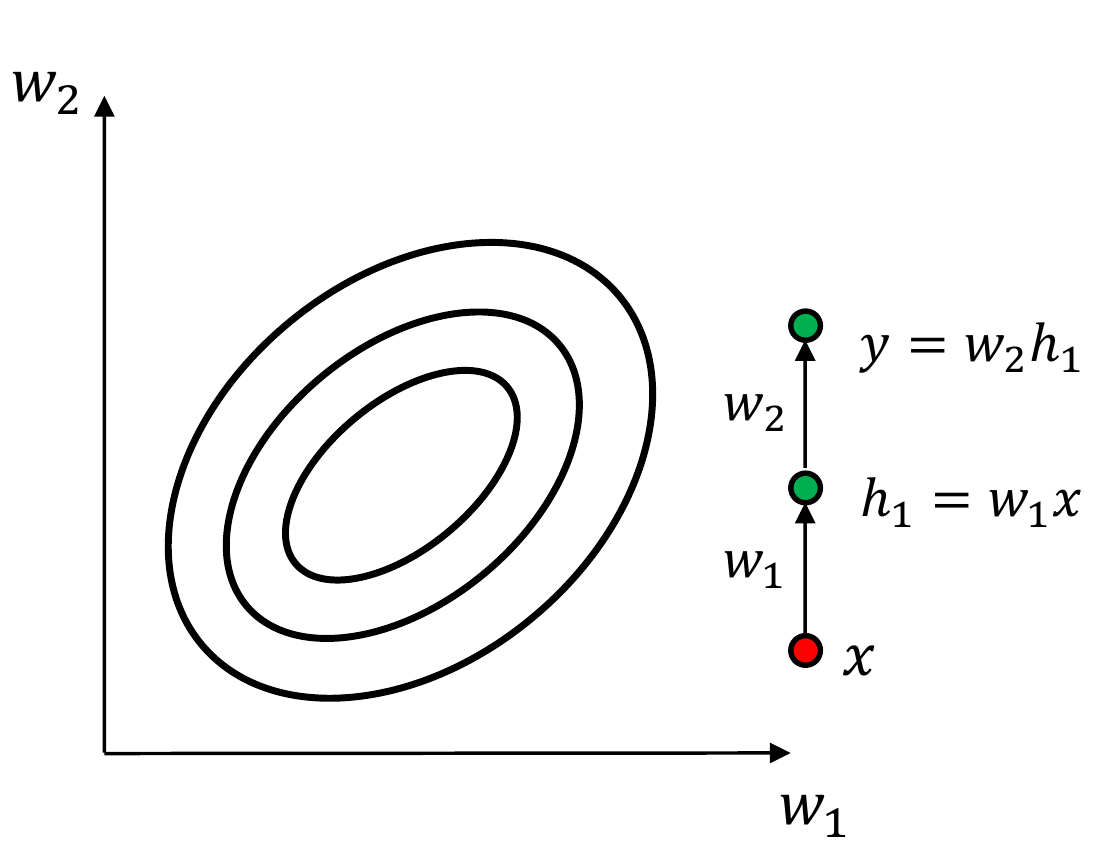}
  }
   \subfigure[scaled parameterizations]{
  \includegraphics[width=0.36\linewidth]{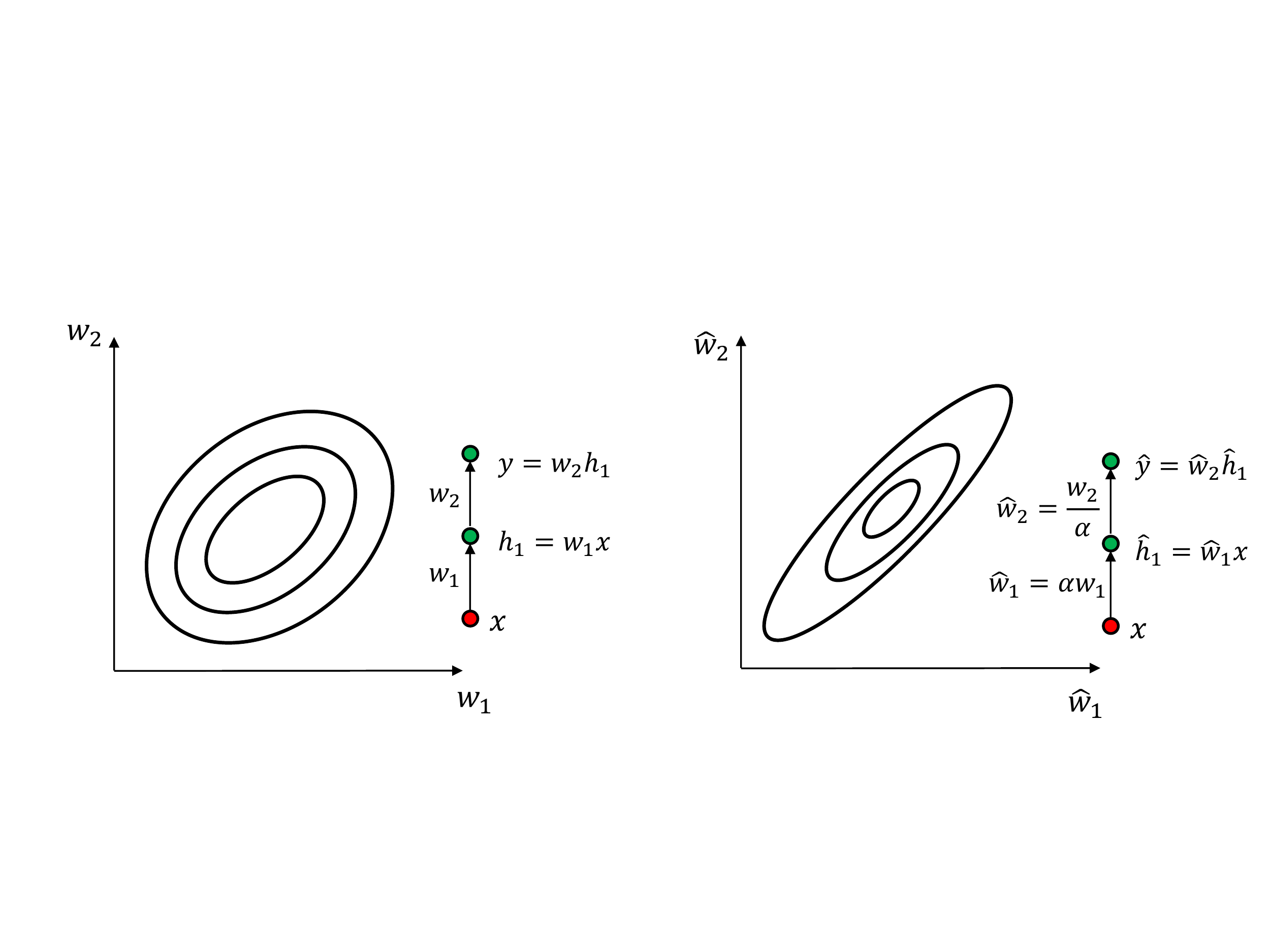}
  }
  \caption{\small An illustrative example of scaling-based weight space symmetry that can cause ill-conditioned problem. (a) the error landscape of $w_1$ and $w_2$ in the same magnitude; (b) the  error landscape of $\hat{w}_1$ and $\hat{w}_2$ by scaling a factor $\alpha$ and $\frac{1}{\alpha}$ respectively in different magnitudes.}
  \label{fig:motivation}
\end{figure*}

For these two models, we can get the back-propagated gradient information $\frac{\partial \mathcal{L} }{\partial y}$ and $\frac{\partial \mathcal{L} }{\partial \hat{y}}$, and further have $\frac{\partial \mathcal{L} }{\partial y}=\frac{\partial \mathcal{L} }{\partial \hat{y}}$ due to the fact $\hat{y}=y$. Based on simple algebra derivation, it is easy to obtain that $\frac{\partial \mathcal{L} }{\partial \hat{w}_2} = \alpha \frac{\partial \mathcal{L} }{\partial w_2}$ and $\frac{\partial \mathcal{L} }{\partial \hat{w}_1} =\frac{1} {\alpha} \frac{\partial \mathcal{L} }{\partial w_1}$. This phenomenon implies that if $w_1$ and $w_2$ are in different magnitude, their gradient information $\frac{\partial \mathcal{L} }{\partial w_1}$ and $\frac{\partial \mathcal{L} }{\partial w_2}$ will be inversely different in terms of magnitude. Subsequently, as $\alpha$ becomes larger, it is more likely that the Hessian matrix will be ill-conditioned, as shown in Figure \ref{fig:motivation}.

\subsection{Formulation for Unit-Norm Constraint}
To relieve the negative effect of \emph{scaling-based weight space symmetry}, in this paper we propose to constrain the incoming weights of each neuron\footnote{We  can  also constrain the outgoing weights to be unit-norm. However, it seems more intuitive with filters being unit-norm.} to be unit-norm. Specifically, we reformulate the optimization problem of Eqn. ~\ref{eqn:optimization_normal} as follows:
 \begin{eqnarray}
\label{eqn:optimization_constrain}
& \min_{\mathbb{W}} ~~\mathbb{E}_{(\mathbf{x},\mathbf{y})\in \mathbb{D}} [\mathcal{L}(\mathbf{y}, f(\mathbf{x}; \mathbb{W}))]  \nonumber \\
& s.t.~~  \text{ddiag}(\mathbf{W}_l \mathbf{W}_l^T)=\mathbf{I}, ~l=1,2,...,L.
\end{eqnarray}
where $\text{ddiag}(\mathbf{M})$ denotes an operation that extracts the diagonal elements of matrix $\mathbf{M}$ and sets the off-diagonal elements as 0. We drop the index of $\mathbf{W}_l$ for simplifying denotation. Indeed, the constraint of the weight matrix $\mathbf{W} \in \mathbb{R}^{n \times p}$ in each layer defines a embedded submanifold of $\mathbb{R}^{n \times p}$ called the Oblique manifold \cite{2006_ICASSP_Absil}:
 \begin{eqnarray}
\mathcal{OB}(n,p)=\{\mathbf{W} \in \mathbb{R}^{n \times p}: \text{ddiag}(\mathbf{W} \mathbf{W}^T)=\mathbf{I}   \}
\end{eqnarray}

Note that here we adopt $\mathcal{OB}(n,p)$ to denote the set of all $n \times p$ matrices with normalized rows, which is  different from the standard denotation with normalized columns \cite{2006_ICASSP_Absil,2008_Book_Absil}.

First, we can apply Riemannian optimization method ~\cite{2008_Book_Absil} to solve Problem~\ref{eqn:optimization_constrain}. We  calculate the Riemannian gradient $\widehat{\frac{\partial \mathcal{L} }{\partial \mathbf{W}}}$ in the tangent space of $\mathcal{OB}(n,p)$ at current point $\mathbf{W}$ by:
 \begin{eqnarray}
 \label{eqn:gradient_Reim}
\widehat{\frac{\partial \mathcal{L} }{\partial \mathbf{W}}}=\frac{\partial \mathcal{L} }{\partial \mathbf{W}} - \text{ddiag}(\mathbf{W} \frac{\partial \mathcal{L} }{\partial \mathbf{W}}^T) \mathbf{W}
\end{eqnarray}
where $\frac{\partial \mathcal{L} }{\partial \mathbf{W}}$ is the ordinary gradient.

Given Riemannian gradient, we update the weight along the negative Riemannian gradient with $-\eta \widehat{\frac{\partial \mathcal{L} }{\partial \mathbf{W}}}$ in the tangent space, where $\eta>0$ is the learning rate. We then use a \emph{retraction} as suggested by \cite{2006_ICASSP_Absil} that maps the tangent vectors to the points on the manifolds as:
\begin{eqnarray}
\label{eqn:retract}
\Upsilon_{\mathbf{W}}(-\eta \widehat{\frac{\partial \mathcal{L} }{\partial \mathbf{W}}})
=(\mathbf{W}-\eta \widehat{\frac{\partial \mathcal{L} }{\partial \mathbf{W}}} )
(\text{ddiag}(\mathbf{M}))^{-1/2}
\end{eqnarray}
where $\mathbf{M}=(\mathbf{W}-\eta \widehat{\frac{\partial \mathcal{L} }{\partial \mathbf{W}}})
(\mathbf{W}-\eta \widehat{\frac{\partial \mathcal{L} }{\partial \mathbf{W}}})^T$. Therefore, we can get the new point $\mathbf{W}^*$ in the Oblique manifold as: $\mathbf{W}^*=\Upsilon_{\mathbf{W}}(-\eta \widehat{\frac{\partial \mathcal{L} }{\partial \mathbf{W}}})$. We update the weight matrices iteratively until convergence.

\section{Projection Based Normalization}
The Riemannian optimization method provides a good solution to Problem \ref{eqn:optimization_constrain}. However, it also introduces extra non-ignorable computational cost. For instance, we have to calculate the Riemannian gradient by subtracting an extra term $\text{ddiag}(\mathbf{W} \frac{\partial \mathcal{L} }{\partial \mathbf{W}}^T) \mathbf{W}$ and then project the weight in the tangent space back to the Oblique manifold by multiplying $(\text{ddiag}(\mathbf{M}))^{-1/2}$ in each iteration. Is it possible to reduce the computational cost without performance loss and meanwhile guarantee the solution satisfying the unit-norm constraints?

To make the following analysis more clear, let us first consider one neuron with its incoming weight $\mathbf{w}$ satisfying the unit-norm constraint $\mathbf{w}^T \mathbf{w}=1$. Based on Eqn. \ref{eqn:gradient_Reim}, its Riemannian gradient $\widehat{\frac{\partial \mathcal{L} }{\partial \mathbf{w}}}$ can be obtained as follows:
 \begin{eqnarray}
 \label{eqn:gradient_Riem_perUnit}
\widehat{\frac{\partial \mathcal{L} }{\partial \mathbf{w}}}=\frac{\partial \mathcal{L} }{\partial \mathbf{w}} - (\mathbf{w}^T \frac{\partial \mathcal{L} }{\partial \mathbf{w}}) \mathbf{w}.
\end{eqnarray}

From Eqn. \ref{eqn:gradient_Riem_perUnit}, we can find that the Riemannian gradient actually adjusts the ordinary gradient by subtracting an extra term $(\mathbf{w}^T \frac{\partial \mathcal{L} }{\partial \mathbf{w}}) \mathbf{w}$. Besides, we have the following fact
\begin{eqnarray}
\| (\mathbf{w}^T \frac{\partial \mathcal{L} }{\partial \mathbf{w}}) \mathbf{w} \| & \leq & |\mathbf{w}^T \frac{\partial \mathcal{L} }{\partial \mathbf{w}}| \| \mathbf{w} \| \nonumber \\
&\leq&  \| \mathbf{w} \| \| \frac{\partial \mathcal{L} }{\partial \mathbf{w}} \| \| \mathbf{w} \|
 = \| \frac{\partial \mathcal{L} }{\partial \mathbf{w}} \|,
\end{eqnarray}
which means that $(\mathbf{w}^T \frac{\partial \mathcal{L} }{\partial \mathbf{w}}) \mathbf{w}$ is not a dominant term compared to $\frac{\partial \mathcal{L} }{\partial \mathbf{w}}$ in Eqn. \ref{eqn:gradient_Riem_perUnit}. We also observe this fact  in our experiments. Therefore, we recommend simply using the ordinary gradient to solve Problem \ref{eqn:optimization_constrain} with much less computation cost as follows:
 \begin{eqnarray}
 \label{eqn:update_norm_perUnit}
   \tilde{\mathbf{w}}=\mathbf{w}- \eta \frac{\partial \mathcal{L} }{\partial \mathbf{w}},\\
  \label{eqn:Norm_projection}
   \mathbf{w}^*= \tilde{\mathbf{w}}/ \| \tilde{\mathbf{w}} \|.
\end{eqnarray}

Here, Eqn. \ref{eqn:Norm_projection} works by projecting the updated weight $\tilde{\mathbf{w}}$ back to the Oblique manifold, and we thus call this operation \emph{norm projection}. Indeed, the operation combining Eqn. \ref{eqn:update_norm_perUnit} and \ref{eqn:Norm_projection} is equivalent to the retractor operation in Eqn. \ref{eqn:gradient_Reim}, when given the Riemannian gradient $\widehat{\frac{\partial \mathcal{L} }{\partial \mathbf{w}}}$.

\begin{figure*}[t]
\centering
\hspace{-0.02\linewidth}
  \subfigure[Effect of norm projection]{
  \includegraphics[width=0.36\linewidth]{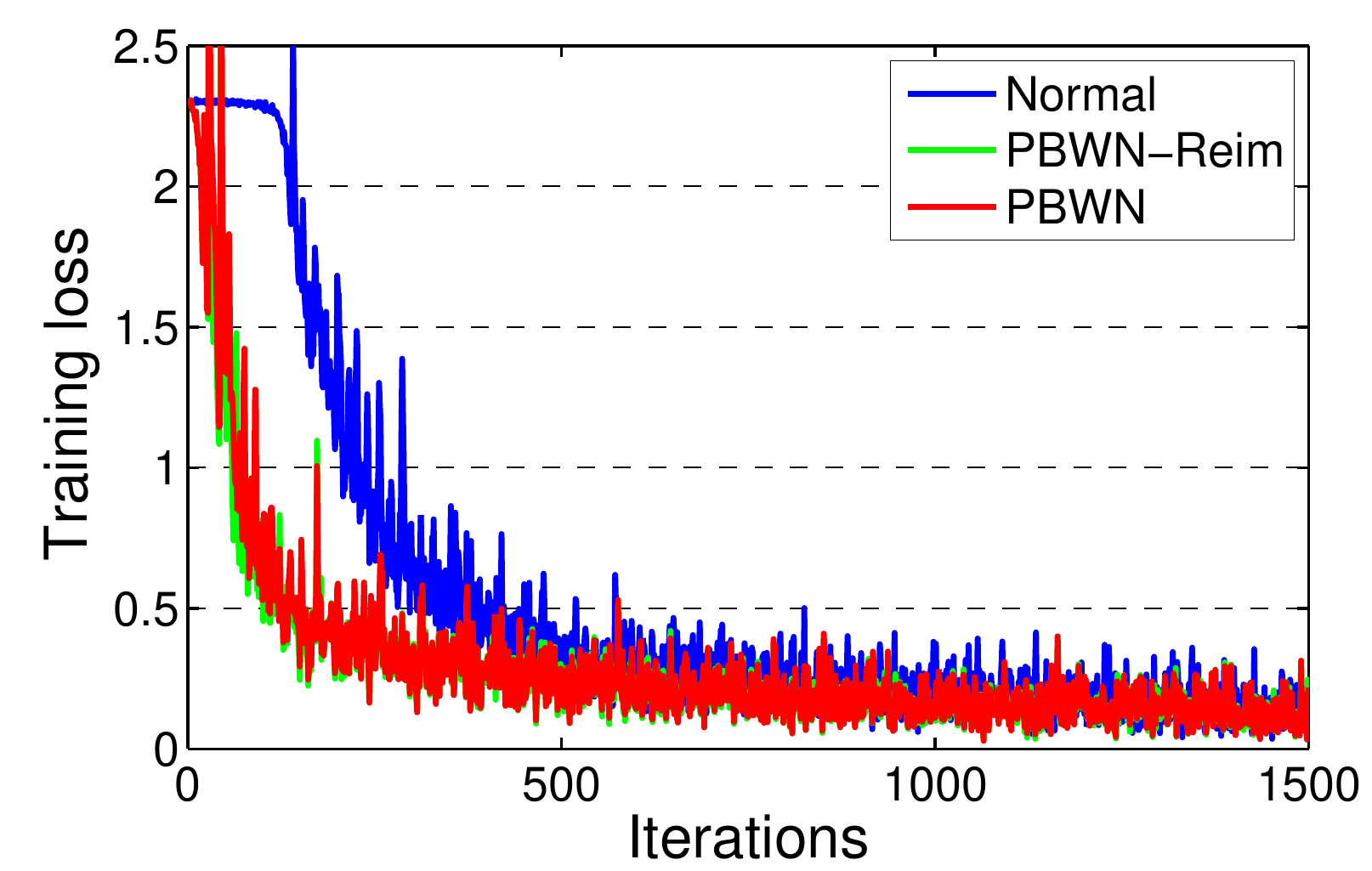}
  }
  \subfigure[Effect of updating intervals]{
  \includegraphics[width=0.36\linewidth]{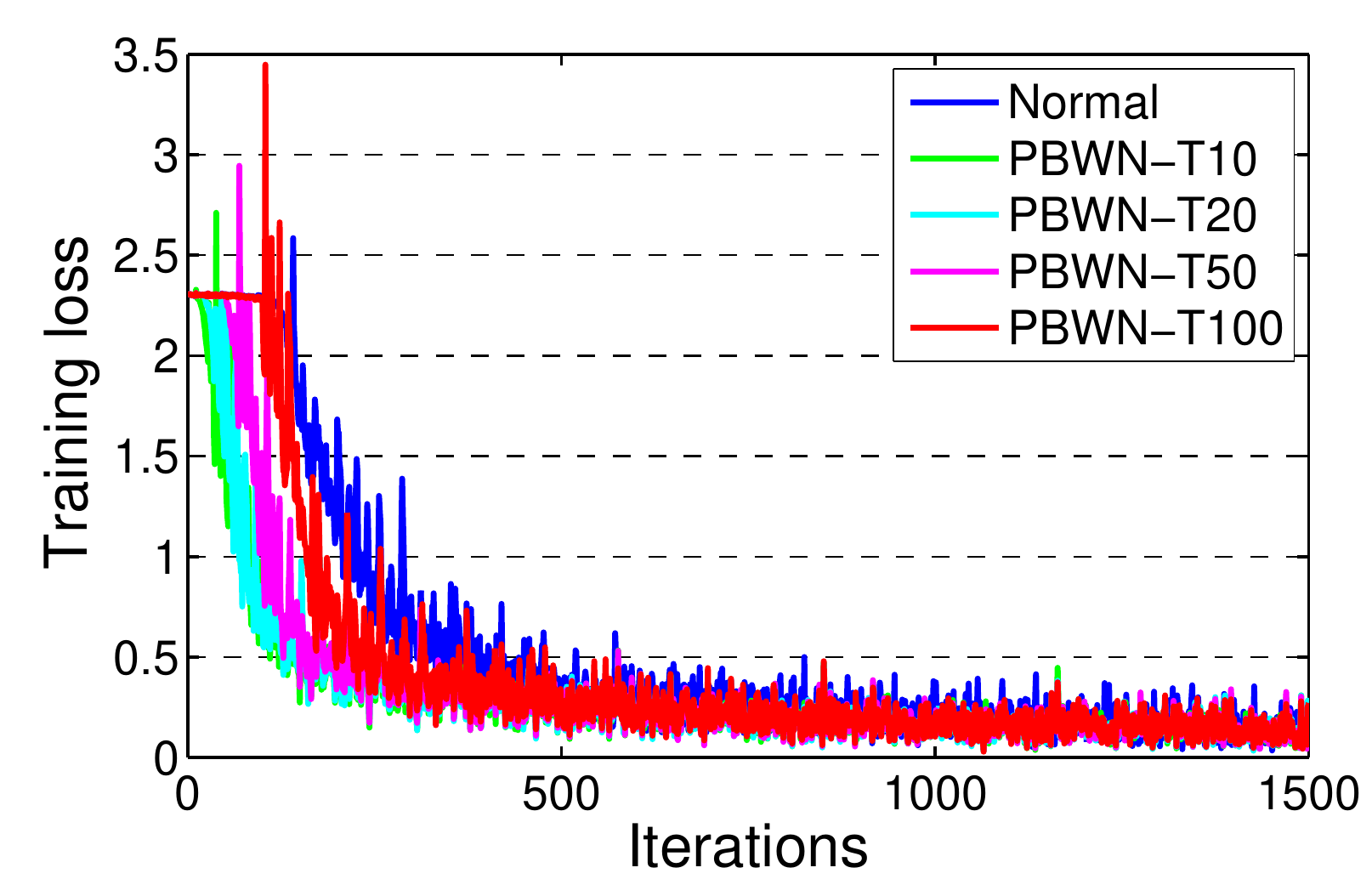}
  }
  \caption{\small An illustrative experiment on MNIST, using multi-layer perceptron (MLP) structure with layer sizes of 1024-750-250-250-10. We train the model by stochastic gradient descent with the mini-batch size of 256. We search the learning rate over $\{0.01,0.03,0.1,0.3,1\}$ and report the best performance of each method (All are under learning rate of 0.3). `Normal' indicates the original network. `PBWN-Riem' and `PBWN' refers to the projection based weight normalization methods that respectively apply \emph{norm projection} for each iteration based on Riemannian and ordinary gradient, while `PBWN-T$T$' performs \emph{norm projection} every $T$ iterations based on ordinary gradient.}
  \label{fig:exp_T}
\end{figure*}

Note that if the weight updating is based on the ordinary gradient in Eqn. \ref{eqn:Norm_projection}, the \emph{norm projection} operation can not make the updating go along the negative gradient direction, and subsequently disturbs the gradient information. We find that such a disturbance eventually does not harm the learning as shown in Figure \ref{fig:exp_T} (a). From it, we observe that using the ordinary gradient has nearly identical training loss curve to using Riemannian gradient.


For more efficient computation, we  can also execute the\emph{ norm projection} operation of Eqn. \ref{eqn:Norm_projection} by an interval $T$ rather than in each iteration. We empirically find this trick can work well in practice. It should be pointed out that when executing \emph{norm projection} operation with a large $T$, our method may lose some information learned in the weight matrix and also suffer instability after the\emph{ norm projection} as shown in Figure \ref{fig:exp_T} (b). From it, we can find that in the initial phase, executing \emph{norm projection} by large interval results in the sudden increase of loss. This is mainly because we change the scale of each filter, which results in the predictions different for the same input. Fortunately, we can remedy this issue by combing with batch normalization \cite{2015_ICML_Ioffe}. We will discuss it in the next subsection.

To summarize, we show our projection  based weight normalization framework in Algorithm \ref{alg_forward}, in which an extra \emph{norm projection} is executed by interval. Note that the proposed Riemannian optimization over Oblique manifold described before can be viewed as a specific instance of our framework, under the conditions that we use Riemannian gradient, steepest gradient descent and interval $T=1$.

\begin{algorithm}[]
   \caption{Projection based weight normalization framework for training DNNs.}
   \label{alg_forward}
  \begin{small}
\begin{algorithmic}[1]

   \STATE \textbf{Input}: A neural network with learnable parameters $\mathbb{W}$, and the updating interval $T$.
     \STATE \textbf{Output}: A trained model with optimal $\mathbb{W}$.
   \STATE Initialize $\mathbb{W}$ by using the regular initialization methods, and set the iteration $t=0$.
    \WHILE {the training is not finished}
    \STATE	Execute forward step to obtain the loss $\mathcal{L}$.
    \STATE	Execute backward step to obtain the gradient information.
    \STATE	Update $\mathbb{W}$ based on the proposed optimization algorithm.
    \STATE  Update the iteration $t\leftarrow t+1$.
     \IF { $mod(t, T)==0$ }
           \STATE Perform \emph{norm projection} on $\mathbf{w} \in \mathbb{W}$ according to (\ref{eqn:Norm_projection}).
     \ENDIF
    \ENDWHILE
\end{algorithmic}
\end{small}
\end{algorithm}


\subsection{Combined with Batch Normalization}
Batch normalization is a popular technique that stabilizes the distribution of activations in each layer and thus accelerates the convergence. It works by normalizing the pre-activation of each neuron to zero-mean and unit-variance over each mini-batch, and an extra learnable scale and bias parameters are recommended to restore the representation power of the networks. Specifically, for each neuron, batch normalization has a formulation as follows:
\begin{eqnarray}
 \label{eqn:BN}
BN(\mathbf{x}; \mathbf{w})= \gamma \frac{\mathbf{w}^T \mathbf{x}- \mathbb{E}(\mathbf{w}^T \mathbf{x})}{\sqrt{Var (\mathbf{w}^T \mathbf{x})}}+\beta.
\end{eqnarray}

One interesting property of batch normalization is that the incoming weight of each neuron is scaling invariant, that is
\begin{eqnarray}
BN(\mathbf{x}; \alpha \mathbf{w})=BN(\mathbf{x}; \mathbf{w}).
\end{eqnarray}
The \emph{norm projection} operation of Eqn. \ref{eqn:Norm_projection} can be viewed as a scaling of $\alpha=\frac{1}{\| \tilde{\mathbf{w}} \|}$. Therefore, when combined with batch normalization, the \emph{norm projection} also can keep the same output during training in a rectifier network, that is $\mathcal{L}(\mathbf{x}; \alpha \mathbf{w})=\mathcal{L}(\mathbf{x}; \mathbf{w})$. Therefore, we can ensure that \emph{norm projection} does not drop any learned information in the weight matrix, even thought we execute \emph{norm projection} outside the gradient descent steps.

Another interesting point is that \emph{norm projection} eventually affects the backpropagation information when combined with batch normalization. Batch normalization owns a property of
\begin{eqnarray}
\frac{\partial BN(\mathbf{x}; \alpha \mathbf{w}) }{\partial (\alpha \mathbf{w})}=\frac{1}{\alpha}   \frac{\partial BN(\mathbf{x}; \mathbf{w}) }{\partial  \mathbf{w}}.
\end{eqnarray}
Therefore, we can get
\begin{eqnarray}
\frac{\partial \mathcal{L} }{\partial (\alpha \mathbf{w})}
=\frac{\partial \mathcal{L} }{\partial BN(\mathbf{x}; \alpha \mathbf{w}) }   \frac{\partial BN(\mathbf{x}; \alpha \mathbf{w})  }{\partial (\alpha \mathbf{w})}
=\frac{1}{\alpha } \frac{\partial \mathcal{L} }{\partial  \mathbf{w}}.
\end{eqnarray}
This indicates that the \emph{norm projection} operation implicitly adjusts the learning rate by a factor of $\| \tilde{\mathbf{w}}\|$.

To summarize, when combined with batch normalization in a rectifier network, the \emph{norm projection} operation  enjoys the following characteristics: (1) guaranteeing that the incoming weight $\mathbf{w}$ is unit-norm; (2) keeping the output same as before the operation during the training; (3) implicitly adjusting the learning rate by a factor of $\| \tilde{\mathbf{w}}\|$. These characteristics make our projection based weight normalization have stable optimization process.

\subsection{Connecting to Weight Decay}
We find that our projection based weight normalization has strong connections to weight decay~\cite{1992_WD_Krogh}. Weight decay ~\cite{1992_WD_Krogh} is a simple yet effective technique to regularize the neural networks. The update formulation of weight decay is:
    \begin{eqnarray}
\label{eqn:WD}
\mathbf{w}^*= \mathbf{w}- \lambda \mathbf{w} - \eta \frac{\partial \mathcal{L} }{\partial \mathbf{w}},
\end{eqnarray}
where $\lambda>0$ is a constant weight decay factor. Indeed, weight decay can be considered as a solution to the loss function $\mathcal{L}(\mathbf{y}, f(\mathbf{x}; \theta))$ appended with a regularization term $\lambda \| \mathbf{w} \|^2$. From this perspective, we can treat weight decay as a soft constraint and while our method a hard constraint with each neuron's incoming weight $\| \mathbf{w}\|=1$.


From another perspective, we can get the weight updating formulation of our method based on Eqn. \ref{eqn:update_norm_perUnit}
and \ref{eqn:Norm_projection}:
\begin{eqnarray}
 \label{eqn:weight}
\mathbf{w}^*= \mathbf{w} - \frac{\lambda_{\eta,\mathbf{w}}-1 }{\lambda_{\eta,\mathbf{w}}} \mathbf{w}
- \frac{\eta}{\lambda_{\eta,\mathbf{w}} }\frac{\partial \mathcal{L} }{\partial \mathbf{w}}
\end{eqnarray}
where  $\lambda_{\eta,\mathbf{w}}= \| \mathbf{w} - \eta \frac{\partial \mathcal{L} }{\partial \mathbf{w}} \|$.
We can find that Eqn. \ref{eqn:weight} has a similar weight updating form as weight decay. Particularly, we have a weight-specific decay rate and also a weight-specific learning rate. Therefore, the solution to optimization over Oblique manifold can be viewed as a regularization method with adaptive regularization factors. Eventually, the weight matrix in $\mathcal{OB}(n,p)$ only has free degree of  $(n-1)\times p$.

\subsection{Computational Cost}
\label{sec:computationCost}
Let's consider a standard linear layer: $\mathbf{y}=\mathbf{W} \mathbf{x}$ with $\mathbf{W} \in \mathbb{R}^{n \times p}$ and a mini-batch input data of size $m$. For each iteration, the computational cost of the standard linear layer (i.e., calculating $\mathbf{y}, \frac{\partial \mathcal{L} }{\partial \mathbf{x}}$ and $\frac{\partial \mathcal{L} }{\partial \mathbf{W}}$) is $6m\times n\times p$ FLOPs. The extra cost for Riemannaian optimization is $6n\times p$ FLOPs. When using our \emph{norm projection} with ordinary gradient, the extra cost is $3n\times p$ FLOPs. Particularly, if we use interval $T$, the extra cost is ${3n\times p}/ {T}$ FLOPs. We can find that the computational cost of \emph{norm projection} with interval update $T$ is negligible to that of the standard linear layer.

For a convolution layer with filters $\mathbf{W}_c \in \mathbb{R}^{n \times p \times F_h \times F_w}$, where $F_h$ and $F_w$ respectively indicate the height and width of the filter, we perform norm propagation over the unrolled $\mathbf{W} \in \mathbb{R}^{n \times p\cdot F_h \cdot F_w}$. Assuming the input feature map with size $h \times w$, the cost of the convolution layer is $6m\times n\times p\times F_h\times  F_w\times  h\times w$ FLOPs. Norm projection with interval updating $T$ has an extra cost of ${3n\times p\times F_h\times  F_w}/{T}$ FLOPs, which is also exactly negligible, compared to the convolution operation.

%

%

\begin{table}[t]
\caption{Comparison of test errors ($\%$) on Inception architecture over CIFAR-10 and CIFAR-100. The results are averaged over five random seeds.}
\label{table:BN-Inception}
\vskip 0.0in
\begin{center}
\begin{small}
\begin{tabular}{lccr}
\hline
Methods & CIFAR-10 & CIFAR-100   \\
\hline
Normal     & 6.48 $\pm$ 0.14    & 25.71 $\pm$ 0.15     \\
WN	  & 6.20 $\pm$ 0.07       &  24.22 $\pm$ 0.53  \\
PBWN-Riem	(ours)  & 5.33 $\pm$ 0.19  & \textbf{22.46 $\pm$ 0.25} \\
PBWN	(ours)  & \textbf{5.22 $\pm$ 0.05}     & 22.70 $\pm$ 0.65 \\
PBWN-Epoch (ours)	& 5.46 $\pm$ 0.22 & 22.83 $\pm$ 0.87 \\
\hline
\end{tabular}
\end{small}
\end{center}
   \vspace{-0.15in}
\end{table}

\begin{table}[t]
\caption{Comparison of test errors ($\%$) on VGG architecture over CIFAR-10 and CIFAR-100 dataset. The results are averaged over five random seeds.}
\label{table:VGG}
\vskip 0.0in
\begin{center}
\begin{small}
\begin{tabular}{lccr}
\hline
Methods & CIFAR-10 & CIFAR-100   \\
\hline
Normal     & 7.23 $\pm$ 0.29 & 27.80 $\pm$ 0.31    \\
WN	  & 7.40 $\pm$ 0.21 &   29.86 $\pm$ 0.38  \\
PBWN-Riem	(ours)  & \textbf{6.23 $\pm$ 0.10} & 27.49 $\pm$ 0.35 \\
PBWN	(ours)  & 6.31 $\pm$ 0.11 & 27.33 $\pm$ 0.21 \\
PBWN-Epoch (ours)	  & 6.27 $\pm$ 0.11 & \textbf{26.91 $\pm$ 0.25} \\
\hline
\end{tabular}
\end{small}
\end{center}
   \vspace{-0.15in}
\end{table}

\section{Experiments}
%
In this section, we first conduct extensive experiments for supervised learning on four widely-used image datasets, i.e., CIFAR-10, CIFAR-100, SVHN and ImageNet, and investigate the performance over various types of CNNs. We also consider semi-supervised learning tasks for permutation invariant MNIST dataset by using Ladder network \cite{2015_NIPS_Rasmus}. For all experiments, we adopt random weight initialization by default as described in \cite{1998_NN_Yann}, unless we specify the weight initialization methods.

\begin{table*}[t]
\caption{Comparison of test errors ($\%$) on residual network with variational layers over CIFAR-10 and the results are averaged over five random seeds. `Res-$L$' indicates residual network with $L$ layers, and `BaseLine*' indicates the results reported in \cite{2015_CVPR_He}, for which res-20, 32, 44, 56 are reported by one run, while res-110 is reported with 5 runs.}
\label{table:resnet1}
\vskip 0.0in
\begin{center}
\begin{small}
\begin{tabular}{l|ccccc}
\toprule
     & Res-20 & Res-32 & Res-44 & Res-56 & Res-110 \\
\hline
BaseLine*  &  8.75 &7.51 &7.17 & 6.97& 6.61 $\pm$ 0.16\\
BaseLine  &  7.94 $\pm$ 0.16 &7.70  $\pm$ 0.26  &7.17 $\pm$ 0.25 & 7.21 $\pm$ 0.25  & 7.09 $\pm$ 0.24\\
WN  &  8.12 $\pm$ 0.18 &7.25  $\pm$ 0.14  &6.86 $\pm$ 0.06 & 7.01 $\pm$ 0.52  & 7.56 $\pm$ 1.11\\
PBWN-Riem (ours) &  8.03 $\pm$ 0.17 &7.18  $\pm$ 0.18  &6.69 $\pm$ 0.15 & 6.42 $\pm$ 0.25  & 6.68 $\pm$ 0.31\\
PBWN (ours) &  8.08 $\pm$ 0.07 &7.09  $\pm$ 0.18  &6.89 $\pm$ 0.17 & 6.48 $\pm$ 0.17  & \textbf{6.27 $\pm$ 0.34}\\
PBWN-Epoch (ours) &  \textbf{7.86 $\pm$ 0.25}  &\textbf{6.99 $\pm$ 0.27 } &\textbf{6.59 $\pm$ 0.17}   & \textbf{6.41 $\pm$ 0.13}    & 6.39 $\pm$ 0.45\\
\bottomrule
\end{tabular}
\end{small}
\end{center}
\vskip -0.15in
\end{table*}

\begin{table*}[t]
\caption{Comparison of test errors ($\%$) on residual network with variational layers over CIFAR-100. The results are averaged over five random seeds.}
\label{table:resnet2}
\vskip 0.0in
\begin{center}
\begin{small}
\begin{tabular}{l|ccccc}
\toprule
     & Res-20 & Res-32 & Res-44 & Res-56 & Res-110 \\
\hline
BaseLine  &  32.28 $\pm$ 0.16 &30.62  $\pm$ 0.35  &29.95 $\pm$ 0.66 & 29.07 $\pm$ 0.40  & 28.79 $\pm$ 0.63\\
WN  &  31.90 $\pm$ 0.45 &30.63  $\pm$ 0.37  &29.57 $\pm$ 0.29 & 29.16 $\pm$ 0.45  & 28.38 $\pm$ 0.99\\
PBWN-Riem (ours) &  31.81 $\pm$ 0.28 &30.12  $\pm$ 0.36  &29.15 $\pm$ 0.18 & \textbf{28.13 $\pm$ 0.49}  & \textbf{27.03 $\pm$ 0.33}\\
PBWN (ours) &  31.99 $\pm$ 0.14 &30.21  $\pm$ 0.20  &29.04 $\pm$ 0.43 & 28.23 $\pm$ 0.31  & 27.16 $\pm$ 0.57\\
PBWN-Epoch (ours)  &  \textbf{31.61 $\pm$ 0.40}  &\textbf{29.85 $\pm$ 0.17 } &\textbf{28.83 $\pm$ 0.09 }  & 28.17 $\pm$ 0.24    &   27.15 $\pm$ 0.58\\
\bottomrule
\end{tabular}
\end{small}
\end{center}
\end{table*}
\subsection{The State-of-the-Art CNNs}
In the following part, we evaluated our method on CIFAR (both CIFAR-10 and CIFAR-100) datasets over the state-of-the-art CNNs, including Inception \cite{2014_CoRR_Szegedy}, VGG \cite{2014_CoRR_Simonyan} and residual network \cite{2015_CVPR_He,2016_CoRR_Zagoruyko}. CIFAR-10 consists of 50,000 training images and 10,000 test images from 10 classes, while CIFAR-100 from 100 classes. Each input image consists of $32\times 32$ pixels. The dataset was preprocessed as described in \cite{2015_CVPR_He} by subtracting the means and dividing the variance for each channel. We follow the simple data augmentation that 4 pixels are padded on each side, and a 32 $\times$ 32 crop is randomly sampled from the padded image or its horizontal flip as described in \cite{2015_CVPR_He}.

We refer to the original networks  as `Normal'.
For our projection based weight normalization methods, we evaluate three setups as follows: (1) `PBWN-Riem': performing \emph{norm projection} for each iteration based on Riemannian gradients; (2) `PBWN': performing \emph{norm projection} for each iteration based on ordinary gradients; (3) `PBWN-Epoch': performing \emph{norm projection} for each epoch based on ordinary gradients. We also choose another very related work named Weight Normalization \cite{2016_CoRR_Salimans} (referred to as `WN') as one baseline.

\subsubsection{Inception Architecture}
\begin{figure*}[t]
\centering
\hspace{-0.02\linewidth}
  \subfigure[CIFAR-10]{
  \includegraphics[width=0.36\linewidth]{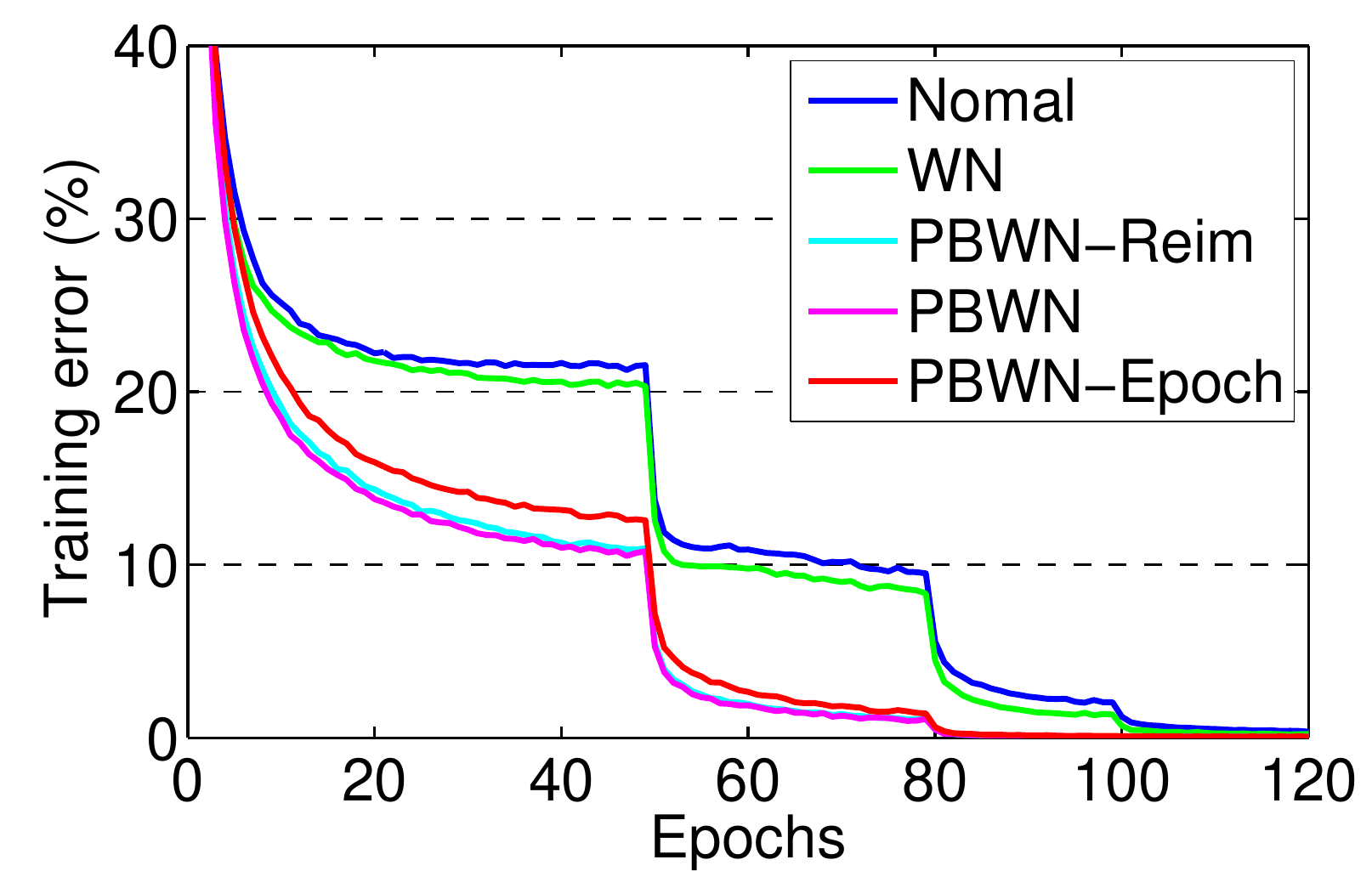}
  }
  \subfigure[CIFAR-100]{
  \includegraphics[width=0.36\linewidth]{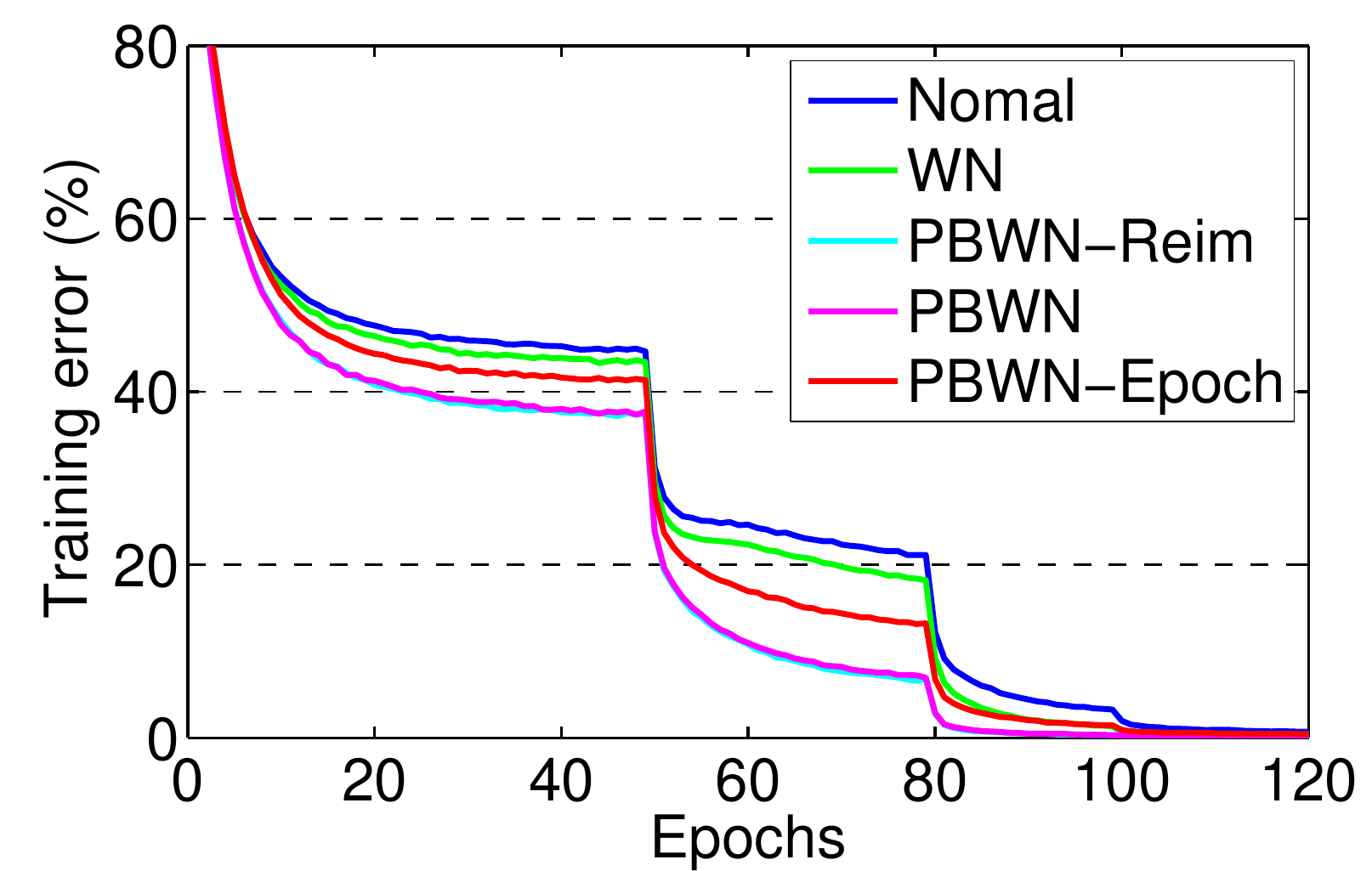}
  }
  \caption{\small Comparison of the training loss with respect to epochs on Inception over CIFAR datasets.}
  \label{fig:exp_Cifar}
    \vspace{-0.15in}
\end{figure*}

We first evaluate our method on Inception architecture~\cite{2014_CoRR_Szegedy} equipped with batch normalization (BN), inserted after each convolution layer. All the models are trained by SGD with a mini-batch size of 64, considering the memory constraints on one GPU. We adopt a momentum of 0.9 and weight decay of 0.0005. Regarding the learning rate annealing, we start with a learning rate of 0.1, divide it by 5 at 50, 80 and 100 epochs, and terminate the training at 120 epochs empirically. The results are also obtained by averaging over five random seeds. Figure \ref{fig:exp_Cifar} (a) and (b) show the training error with respect to epochs on CIFAR-10 and CIFAR-100 dataset respectively, and Table \ref{table:BN-Inception} lists the test errors. From Figure \ref{fig:exp_Cifar}, we observe that our model can converge significantly faster than the baselines. Particularly, `PBWN-Riem' and `PBWN' have nearly identical training curves, which means that there is no need to calculate the Reimannian gradient when performing \emph{norm projection} in Inception network with BN. The test performance in Table \ref{table:BN-Inception} further demonstrates that our methods also can achieve significant improvements over the baselines, mainly owing to their desirable regularization ability.

\subsubsection{VGG Architecture}
We further investigate the performance on the VGG-E architecture \cite{2014_CoRR_Simonyan} with global average pooling and batch normalization inserted after each convolution layer. We initialize the model with \emph{He-Init} \cite{2015_ICCV_He}. The models are again trained by SGD with a mini-batch size of 128, the momentum of 0.9 and weight decay of 0.0005. Here, we start with a learning rate of 0.1, divide it by 5 at 80 and 120 epochs, and terminate the training at 160 epochs empirically. The averaged test errors after training are shown in Table \ref{table:VGG}, from which we can easily get the same conclusion as Inception architecture that our model can significantly boost the test performance of the baselines.

\subsubsection{Residual Network}
In this experiment, we further apply our method on famous residual network architecture \cite{2015_CVPR_He}. We follow the exactly same experimental protocol as described in \cite{2015_CVPR_He} and adopt the publicly available Torch implementation\footnote{https://github.com/facebook/fb.resnet.torch} for residual network. Table \ref{table:resnet1} and \ref{table:resnet2} respectively show all the results of different methods on CIFAR-10 and CIFAR-100, using the residual network architecture with varied depths $L=\{20, 32, 44, 56, 110 \}$. We can find that our methods consistently achieve better performance when using different depths. Especially, with the depth increasing, our methods obtain more performance gains. Besides, we observe that there is no significant difference among the performance of different \emph{norm projection} methods, when using different gradient information or updating intervals. Indeed, `PBWN-Epoch' works the best for most cases.  This further indicates the effectiveness of our efficient model by executing norm projection by interval, meanwhile without performance degeneration.

\subsubsection{Efficiency Analysis}
We also investigate the wall clock times of training above networks, including Inception, VGG and 110 layer residual network. The experiment is implemented based on Torch and conducted on one Tesla K80 GPU. From the results reported in Table \ref{table:TimeCost}, we can find that our `PBWN-epoch' costs almost the same time as `Normal' on all architectures, which means that it does not introduce extra time cost in practice as we analyzed in previous sections. `PBWN' also requires little extra time cost, while `PBWN-Riem' needs non-ignorable extra time. The results show that the \emph{norm projection} solution can faithfully improve the efficiency of the optimization with unit-norm constraints and meanwhile achieve satisfying performance.

\subsection{Large-Scale Classification Task}
\paragraph{SVHN dataset}
To comprehensively study the performance of the proposed method, we consider a larger datasets SVHN \cite{2011_NIPS_Netzer} for digit recognition. SVHN consists of $32 \times 32$ color images of house numbers collected by Google Street View. It includes 73,257 train images and 26,032 test images. Besides, we further appended the extra augmented 531,131 images into the training set. The experiment is based on wide residual network that achieves the state-of-the-art results on this dataset. We use the WRN-16-4 as \cite{2016_CoRR_Zagoruyko} does, and follow the experimental setting provided in \cite{2016_CoRR_Zagoruyko}: (1) The input images are divided by 255 to ensure them in [0,1] range; (2) During the training, SGD is used with momentum of 0.9 and dampening to 0, weight decay of 0.0005 and mini-batch size of 128. The initial learning rate is set to 0.01 and dropped at 80 and 120 epochs by 0.1, until the total 160 epochs complete. Dropout is set to 0.4. Here, we only apply our method `PBWN-Epoch' on this WRN-16-4 architecture, namely, we execute \emph{norm projection} per epoch considering the time cost for such a large dataset. The results are shown in Table \ref{table:WR} comparing several state-of-the-art methods in the literature. It can be easily to see that WRN achieves the best performance compared to other baselines, and our method can further improves WRN by simply executing the efficient \emph{norm projection} operation for each epoch.
\begin{table}[t]
\caption{Time costs (hour) of different methods spent on training Inception, VGG and 110 layer residual networks.}
\label{table:TimeCost}
\vskip 0.0in
\begin{center}
\begin{small}
\begin{tabular}{lcccr}
\hline
Methods   & Inception & VGG & Res-110   \\
\hline
Normal            &  20.96   &  4.20   & 5.96    \\
WN	             &  23.33   &  5.27  &  6.42  \\
PBWN-Riem &   23.92  &  5.01  &  7.49 \\
PBWN	       &  21.21   &  4.23  &  6.29 \\
PBWN-Epoch 	 &  20.97  &  4.20  &   5.97 \\
\hline
\end{tabular}
\end{small}
\end{center}
  \vspace{-0.15in}
\end{table}

\begin{table}[t]
\caption{Comparison of test errors ($\%$) on SVHN dataset. WRN* indicates our reproduced results.}
\label{table:WR}
\vskip 0.0in
\begin{center}
\begin{small}
\begin{tabular}{lcc|cr}
\hline
Methods & test error   \\
\hline
DSN~\cite{2015_AISTATS_Lee}     & 1.92 \\
RSD ~\cite{2016_ECCV_Huang}    & 1.75 \\
GPF ~\cite{2015_AISTATS_Lee}   & 1.69 \\
WRN  ~\cite{2016_CoRR_Zagoruyko}   & 1.64 \\
\hline
WRN*    & 1.644($\pm$ 0.046) \\
WRN-PBWN-Epoch    & \textbf{1.607}($\pm$ 0.005) \\
\hline
\end{tabular}
\end{small}
\end{center}
  \vspace{-0.1in}
\end{table}


\begin{table}[t]
  \caption{Comparison of test errors ($\%$) on 34 layers residual networks and its pre-activation version over ImageNet-2012 dataset.}
  \label{table:ImageNet}
  \vspace{0.1in}
  \centering
  \begin{small}
  \begin{tabular}{c|cc|cc}
    \toprule
    & \multicolumn{2}{c|}{Residual} &    \multicolumn{2}{c}{Pre-Residual}   \\
    method   & Top-1    & Top-5  & Top-1   & Top-5 \\
  \hline
     Normal    & 28.62       &  9.69     &     28.81       & 9.78  \\
     PBWN-Epoch     & \textbf{27.88} & \textbf{9.23}   &\textbf{28.2 } & \textbf{9.45} \\
    \bottomrule
  \end{tabular}
\end{small}
  \vspace{-0.1in}
\end{table}

\begin{table*}[t]
\caption{Comparison of test errors ($\%$) for semi-supervised setup on permutation invariant MNIST dataset. We show the test error for a given number of samples=$\{20, 50, 100\}$. Ladder* indicates  our implementation of Ladder network \cite{2015_NIPS_Rasmus}.}
\label{table:semi}
\vskip 0.0in
\begin{center}
\begin{small}
\begin{tabular}{l|ccc}
\toprule
method & \multicolumn{3}{c}{Test error($\%$) for a given number of labeled samples} \\
     & 20 & 50 & 100  \\
\hline

CatGAN \cite{2016_ICLR_Springenberg} &  -  & - & 1.91 $\pm$ 0.1   \\
Skip Deep Generative Model \cite{2016_ICML_Maal} & - & - &  1.32 $\pm$ 0.07   \\
Auxiliary Deep Generative Model\cite{2016_ICML_Maal} & - & - &  0.96 $\pm$ 0.02     \\
Virtual Adversarial \cite{2017_CoRR_Miyato} &  - &- & 1.36  \\
Ladder \cite{2015_NIPS_Rasmus}& - & 1.62 $\pm$ 0.65 & 1.06 $\pm$ 0.37    \\
Ladder+AMLP \cite{2016_ICML_Pezeshki}& - &-  &  1.002 $\pm$ 0.038  \\
GAN with feature matching \cite{2016_NIPS_Goodfellow}& 16.77 $\pm$ 4.52 &  2.21 $\pm$ 1.36 & 0.93  $\pm$ 0.065  \\
Triple-GAN \cite{2017_Corr_Li}& 4.81 $\pm$ 4.95 &  1.56 $\pm$ 0.72 & \textbf{0.91  $\pm$ 0.58}  \\
\hline
Ladder* (our implementation) & 9.67 $\pm$ 10.1 & 3.53 $\pm$ 6.6 &  1.12 $\pm$ 0.59  \\
Ladder+PBWN (ours) & \textbf{2.52 $\pm$ 2.42 }& \textbf{1.06 $\pm$ 0.48} & \textbf{0.91  $\pm$ 0.05}  \\
\hline
\bottomrule
\end{tabular}
\end{small}
\end{center}
  \vspace{-0.1in}
\end{table*}

\paragraph{ImageNet 2012}
To further validate the effectiveness of our method on large-scale dataset, we employ ImageNet 2012 consisting of 1,000 classes~\cite{2009_ImageNet}. We train the models on the given official 1.28M  training images, and evaluated on the validation set with 50k images. We evaluate the classification performance based on top-1 and top-5 error. Note that in this part, we mainly focus on whether our proposed method is able to handle diverse and large-scale datasets and provide a relative benefit for the conventional architecture, rather than achieving the state-of-the-art results. We use the 34 layers residual network \cite{2015_CVPR_He} and its pre-activation version ~\cite{2016_CoRR_He} to perform the classification task. The stochastic gradient descent is again applied with a mini-batch size of 64, a momentum of 0.9 and a weight decay of 0.0001. We use exponential decay to $1/100$ of the initial learning rate until the end of 50 training epochs.  We run with the initial learning rate of $\{0.05, 0.1\}$ and select the best results shown in Table \ref{table:ImageNet}. We can find that `PBWN-Epoch' achieves lower test errors compared to the original residual network and pre-activation residual networks.

\subsection{Semi-supervised Learning for Permutation Invariant MNIST }
In this section, we applied our proposed method to semi-supervised learning tasks on Ladder network \cite{2015_NIPS_Rasmus} over the permutation invariant MNIST dataset. Three semi-supervised classification tasks are considered respectively with 20, 50, 100 labeled examples. These labeled examples are sampled randomly with a balanced number for each class.

We re-implement Ladder network based on Torch, following the Theano implementation by \cite{2015_NIPS_Rasmus}. Specifically, we adopt the setup as described in \cite{2015_NIPS_Rasmus} and \cite{2016_ICML_Pezeshki}: (1) the layer sizes of the model is 784-1000-500-250-250-250-10; (2) the models are trained by Adam optimization \cite{2014_CoRR_Kingma} respectively with mini-batch size of 100 (the task of 100  labeled examples), 50 (the task of 50 labeled examples) and 20 (the task of 20 labeled examples); (3) all the models are trained for 50,000 iterations with the initial learning rate, followed by 25,000 iterations with a decaying linearly to 0. We execute simple hyper-parameters search with learning rate in $\{0.002, 0.001, 0.0005 \}$ and weight decay in $\{0.0002, 0.0001, 0\}$\footnote{The detailed experimental configuration  to reproduce our results in our codes available on: https://github.com/huangleiBuaa/NormProjection}. In this case, all experiments are run with 10 random seeds.

In Table \ref{table:semi}, we report the results of Ladder based on our implementation (denoted by Ladder*) and our `PBWN' that performs \emph{norm projection} in each iteration.
 From Table \ref{table:semi}, we can see that our method significantly improves the performance of the original Ladder network and achieves new state-of-the-art results in the tasks with 20, 50, and 100 labeled examples. Especially, with 20 labeled examples our method achieves $2.52\%$ test error. We conjecture that these appealing results of our method are mainly stemming from its well regularization ability. 

\section{Related Work and Discussion}
There exist a number of methods that regularize neural networks by bounding the magnitude of weights. One commonly used method is weight decay ~\cite{1992_WD_Krogh}, which can be considered as a solution to the loss function  appended with a regularization term of squared \emph{L2-norm }of the weight vector.
Max-norm ~\cite{2005_Nathan_2005,2014_JMLR_Nitish} constrains the norm of the incoming weights at each hidden unit to be bounded by a constant. It can be viewed as a constrained optimization problem over a ball in the parameter space, while our method addresses the optimization problem over an Oblique manifold. Path normalization ~\cite{2015_NIPS_Neyshabur} follows the idea of max-norm, but bounds the product of weights along a path from the input to output nodes, which can also be viewed as a regularizer as weight decay \cite{1992_WD_Krogh}. Weight normalization ~\cite{2016_CoRR_Salimans} decouples the length of each incoming weight vector from its directions. If the extra scaling parameter is not considered, weight normalization  can  be viewed as normalizing the incoming weight. However, it solves the problem via re-parameterization and can not guarantee whether the conditioning of Hessian matrix over proxy parameter will be improved; while our method performs normalization via projection and optimization over the original parameter space, which ensures the improvement of conditioning of Hessian matrix as shown in Figure~\ref{fig:motivation}. We experimentally show that our method outperforms weight normalization~\cite{2016_CoRR_Salimans}, in terms of both the effectiveness and computation efficiency.

There are large amount of work introducing orthogonality to the weight matrix ~\cite{2016_ICML_Arjovsky,2016_NIPS_Wisdom,2016_CoRR_Dorobantu,2017_ICML_Eugene,2017_Corr_Harandi,2016_Corr_Ozay,Huang_2017_arxiv} in deep neural networks to address the gradient vanish and explosion problem. Solving the problem with such orthogonality constraint is usually limited to the hidden-to-hidden transformation in Recurrent neural networks~\cite{2016_ICML_Arjovsky,2016_NIPS_Wisdom,2016_CoRR_Dorobantu,2017_ICML_Eugene}. Some work also consider orthogonal weight matrix in feed forward neural networks~\cite{2017_Corr_Harandi,2016_Corr_Ozay,Huang_2017_arxiv}, while their solutions introduce expensive computation costs.

Normalizing the activations ~\cite{2015_ICML_Ioffe,2016_CoRR_Ba,2017_ICLR_Ren} in deep neural networks have also been studied. Batch normalization~\cite{2015_ICML_Ioffe} is a famous and effective technique to normalize the activations. It standardizes the  pre-activation of each neuron to zero-mean and unit-variance over each mini-batch. Layer normalization ~\cite{2016_CoRR_Ba} computed the  statics of zero-mean and unit-variance over all the hidden units in the same layers, targeting at the scenario where the size of mini-batch is limited. Division normalization ~\cite{2017_ICLR_Ren} is proposed from a unified view of normalization, which includes batch and layer normalization as special cases. These methods focus on normalizing the activations and are data dependent normalization, while our method normalizing the weights and therefore is data independent normalization. Based on the fact that our method is orthogonal to these methods, we provide analysis and experimental results showing that our method can improve the performance of batch normalization by combining them together.

Concurrent to our work, Cho and Lee ~\cite{2017_Corr_Cho} propose to optimize over Grassmann manifold,  aiming to improve the performance of neural networks equipped with batch normalization~\cite{2015_ICML_Ioffe}.
The differences between their work and our work are in two aspects: (1) they only use the traditional Riemannian optimization method (`Riemannian gradient + exponential maps'~\cite{2004_Math}) to solve the constraint optimization problem, which introduce non-trivial commutation cost; while we consider both Riemannian optimization method (`Riemannian gradient+ retraction' ~\cite{2006_ICASSP_Absil} ) and further proposed a more general and efficient projection based weight normalization framework, which introduces negligible extra computation cost; (2) ~\cite{2017_Corr_Cho} requires gradient clipping technique~\cite{2013_ICML_Pascanu} to make optimization stable and also needs tailored revision for SGD with momentum. On the contrary, our method is more general without requiring any extra tailored revision, and it can also collaborate well with other techniques of training neural networks.

\section{Conclusions}
The scaling-based weight space symmetry can cause ill-conditioning problem when optimizing deep neural networks. In this paper, we propose to address the problem by constraining the incoming weights of each neuron to be unit-norm. We provide the projection based weight normalization method, which serves as a simple, yet effective and efficient solution to such a constrained optimization problem. Our extensive experiments demonstrate that the proposed method greatly improves the performance of various state-of-the-art network architectures over large scale datasets. We show that the projection based weight normalization offers a good direction for improving the performance of deep neural networks by alleviating the ill-conditioning problem.

\bibliographystyle{plain}
\bibliography{whitening_PN}

\begin{thebibliography}{10}

\bibitem{2006_ICASSP_Absil}
P.~A. Absil and K.~A. Gallivan.
\newblock Joint diagonalization on the oblique manifold for independent
  component analysis.
\newblock In {\em ICASSP}, 2006.

\bibitem{2004_Math}
P.-A. Absil, R.~Mahony, and R.~Sepulchre.
\newblock Riemannian geometry of {G}rassmann manifolds with a view on
  algorithmic computation.
\newblock {\em Acta Appl. Math.}, 80(2):199--220, 2004.

\bibitem{2008_Book_Absil}
P.-A. Absil, R.~Mahony, and R.~Sepulchre.
\newblock {\em Optimization Algorithms on Matrix Manifolds}.
\newblock Princeton University Press, Princeton, NJ, 2008.

\bibitem{2016_ICML_Arjovsky}
Mart{\'{\i}}n Arjovsky, Amar Shah, and Yoshua Bengio.
\newblock Unitary evolution recurrent neural networks.
\newblock In {\em ICML}, 2016.

\bibitem{2016_CoRR_Ba}
Lei~Jimmy Ba, Ryan Kiros, and Geoffrey~E. Hinton.
\newblock Layer normalization.
\newblock {\em CoRR}, abs/1607.06450, 2016.

\bibitem{1993_NC_Chen}
An~Mei Chen, Haw minn Lu, and Robert Hecht-Nielsen.
\newblock On the geometry of feedforward neural network error surfaces.
\newblock {\em Neural Computation}, 5(6):910--927, 1993.

\bibitem{2017_Corr_Cho}
Minhyung Cho and Jaehyung Lee.
\newblock Riemannian approach to batch normalization.
\newblock {\em CoRR}, abs/1709.09603, 2017.

\bibitem{2009_ImageNet}
J.~Deng, W.~Dong, R.~Socher, L.-J. Li, K.~Li, and L.~Fei-Fei.
\newblock {ImageNet: A Large-Scale Hierarchical Image Database}.
\newblock In {\em CVPR}, 2009.

\bibitem{2016_CoRR_Dorobantu}
Victor Dorobantu, Per~Andre Stromhaug, and Jess Renteria.
\newblock Dizzyrnn: Reparameterizing recurrent neural networks for
  norm-preserving backpropagation.
\newblock {\em CoRR}, abs/1612.04035, 2016.

\bibitem{2010_AISTATS_Glorot}
Xavier Glorot and Yoshua Bengio.
\newblock Understanding the difficulty of training deep feedforward neural
  networks.
\newblock In {\em AISTATS}, 2010.

\bibitem{Goodfellow-et-al-2016}
Ian Goodfellow, Yoshua Bengio, and Aaron Courville.
\newblock {\em Deep Learning}.
\newblock MIT Press, 2016.

\bibitem{Goodfellow_CoRR_2013}
Ian~J. Goodfellow, David Warde{-}Farley, Mehdi Mirza, Aaron~C. Courville, and
  Yoshua Bengio.
\newblock Maxout networks.
\newblock In {\em ICML}, 2013.

\bibitem{2015_ICML_Grosse}
Roger~B. Grosse and Ruslan Salakhutdinov.
\newblock Scaling up natural gradient by sparsely factorizing the inverse
  fisher matrix.
\newblock In {\em {ICML}}, 2015.

\bibitem{2017_Corr_Harandi}
Mehrtash Harandi and Basura Fernando.
\newblock Generalized backpropagation, etude de cas: Orthogonality.
\newblock In {\em arxiv}, 2017.

\bibitem{2015_ICCV_He}
Kaiming He, Xiangyu Zhang, Shaoqing Ren, and Jian Sun.
\newblock Delving deep into rectifiers: Surpassing human-level performance on
  imagenet classification.
\newblock In {\em {ICCV}}, 2015.

\bibitem{2015_CVPR_He}
Kaiming He, Xiangyu Zhang, Shaoqing Ren, and Jian Sun.
\newblock Deep residual learning for image recognition.
\newblock In {\em CVPR}, 2016.

\bibitem{2016_CoRR_He}
Kaiming He, Xiangyu Zhang, Shaoqing Ren, and Jian Sun.
\newblock Identity mappings in deep residual networks.
\newblock {\em CoRR}, abs/1603.05027, 2016.

\bibitem{2016_ECCV_Huang}
Gao Huang, Yu~Sun, Zhuang Liu, Daniel Sedra, and Kilian~Q. Weinberger.
\newblock Deep networks with stochastic depth.
\newblock In {\em ECCV}, pages 646--661, 2016.

\bibitem{Huang_2017_arxiv}
Lei Huang, Xianglong Liu, Bo~Lang, Admas~Wei Yu, and Bo~Li.
\newblock Orthogonal weight normalization: Solution to optimization over
  multiple dependent stiefel manifolds in deep neural networks.
\newblock {\em CoRR}, abs/1709.06079, 2017.

\bibitem{2015_ICML_Ioffe}
Sergey Ioffe and Christian Szegedy.
\newblock Batch normalization: Accelerating deep network training by reducing
  internal covariate shift.
\newblock In {\em ICML}, 2015.

\bibitem{2014_CoRR_Kingma}
Diederik~P. Kingma and Jimmy Ba.
\newblock Adam: {A} method for stochastic optimization.
\newblock {\em CoRR}, abs/1412.6980, 2014.

\bibitem{2009_TR_Alex}
Alex Krizhevsky.
\newblock Learning multiple layers of features from tiny images.
\newblock Technical report, 2009.

\bibitem{1992_WD_Krogh}
Anders Krogh and John~A. Hertz.
\newblock A simple weight decay can improve generalization.
\newblock In {\em NIPS}. 1992.

\bibitem{1998_NN_Yann}
Yann LeCun, L{\'e}on Bottou, Genevieve~B. Orr, and Klaus-Robert M\"{u}ller.
\newblock Effiicient backprop.
\newblock In {\em Neural Networks: Tricks of the Trade}, 1998.

\bibitem{2015_AISTATS_Lee}
Chen-Yu Lee, Saining Xie, Patrick~W. Gallagher, Zhengyou Zhang, and Zhuowen Tu.
\newblock Deeply-supervised nets.
\newblock In {\em AISTATS}, volume~38 of {\em JMLR Proceedings}. JMLR.org,
  2015.

\bibitem{2017_Corr_Li}
Chongxuan Li, Kun Xu, Jun Zhu, and Bo~Zhang.
\newblock Triple generative adversarial nets.
\newblock {\em CoRR}, abs/1703.02291, 2017.

\bibitem{2016_ICML_Maal}
Lars Maale, Casper~Kaae Snderby, Sren~Kaae Snderby, and Ole Winther.
\newblock Auxiliary deep generative models.
\newblock In {\em ICML}, 2016.

\bibitem{2017_CoRR_Miyato}
Takeru Miyato, Shin{-}ichi Maeda, Masanori Koyama, and Shin Ishii.
\newblock Virtual adversarial training: a regularization method for supervised
  and semi-supervised learning.
\newblock {\em CoRR}, abs/1704.03976, 2017.

\bibitem{2010_ICML_Nair}
Vinod Nair and Geoffrey~E. Hinton.
\newblock Rectified linear units improve restricted boltzmann machines.
\newblock In {\em {ICML}}, 2010.

\bibitem{2011_NIPS_Netzer}
Yuval Netzer, Tao Wang, Adam Coates, Alessandro Bissacco, Bo~Wu, and Andrew~Y.
  Ng.
\newblock Reading digits in natural images with unsupervised feature learning.
\newblock In {\em NIPS Workshop}, 2011.

\bibitem{2015_NIPS_Neyshabur}
Behnam Neyshabur, Ruslan Salakhutdinov, and Nathan Srebro.
\newblock Path-sgd: Path-normalized optimization in deep neural networks.
\newblock In {\em NIPS}, 2015.

\bibitem{2016_Corr_Ozay}
Mete Ozay and Takayuki Okatani.
\newblock Optimization on submanifolds of convolution kernels in cnns.
\newblock {\em CoRR}, abs/1610.07008, 2016.

\bibitem{2013_ICML_Pascanu}
Razvan Pascanu, Tomas Mikolov, and Yoshua Bengio.
\newblock On the difficulty of training recurrent neural networks.
\newblock In {\em ICML}, 2013.

\bibitem{2016_ICML_Pezeshki}
Mohammad Pezeshki, Linxi Fan, Philemon Brakel, Aaron~C. Courville, and Yoshua
  Bengio.
\newblock Deconstructing the ladder network architecture.
\newblock In {\em ICML}, 2016.

\bibitem{2015_NIPS_Rasmus}
Antti Rasmus, Harri Valpola, Mikko Honkala, Mathias Berglund, and Tapani Raiko.
\newblock Semi-supervised learning with ladder networks.
\newblock In {\em NIPS}, 2015.

\bibitem{2017_ICLR_Ren}
Mengye Ren, Renjie Liao, Raquel Urtasun, Fabian~H. Sinz, and Richard~S. Zemel.
\newblock Normalizing the normalizers: Comparing and extending network
  normalization schemes.
\newblock In {\em ICLR}, 2017.

\bibitem{2016_NIPS_Goodfellow}
Tim Salimans, Ian~J. Goodfellow, Wojciech Zaremba, Vicki Cheung, Alec Radford,
  and Xi~Chen.
\newblock Improved techniques for training gans.
\newblock In {\em NIPS}, pages 2226--2234, 2016.

\bibitem{2016_CoRR_Salimans}
Tim Salimans and Diederik~P. Kingma.
\newblock Weight normalization: {A} simple reparameterization to accelerate
  training of deep neural networks.
\newblock In {\em {NIPS}}, 2016.

\bibitem{2014_CoRR_Simonyan}
Karen Simonyan and Andrew Zisserman.
\newblock Very deep convolutional networks for large-scale image recognition.
\newblock {\em CoRR}, abs/1409.1556, 2014.

\bibitem{2016_ICLR_Springenberg}
Jost~Tobias Springenberg.
\newblock Unsupervised and semi-supervised learning with categorical generative
  adversarial networks.
\newblock In {\em ICLR}, 2016.

\bibitem{2005_Nathan_2005}
Nathan Srebro and Adi Shraibman.
\newblock Rank, trace-norm and max-norm.
\newblock In {\em COLT}, 2005.

\bibitem{2014_JMLR_Nitish}
Nitish Srivastava, Geoffrey Hinton, Alex Krizhevsky, Ilya Sutskever, and Ruslan
  Salakhutdinov.
\newblock Dropout: A simple way to prevent neural networks from overfitting.
\newblock {\em J. Mach. Learn. Res.}, 15(1):1929--1958, January 2014.

\bibitem{2014_CoRR_Szegedy}
Christian Szegedy, Wei Liu, Yangqing Jia, Pierre Sermanet, Scott Reed, Dragomir
  Anguelov, Dumitru Erhan, Vincent Vanhoucke, and Andrew Rabinovich.
\newblock Going deeper with convolutions.
\newblock In {\em CVPR}, 2015.

\bibitem{2017_AAAI_Tu}
Zhaopeng Tu, Yang Liu, Lifeng Shang, Xiaohua Liu, and Hang Li.
\newblock Neural machine translation with reconstruction.
\newblock In {\em AAAI}, 2017.

\bibitem{2017_ICML_Eugene}
Eugene Vorontsov, Chiheb Trabelsi, Samuel Kadoury, and Chris Pal.
\newblock On orthogonality and learning recurrent networks with long term
  dependencies.
\newblock In {\em ICML}, 2017.

\bibitem{2014_ICASSP_Wiesler}
Simon Wiesler, Alexander Richard, Ralf Schl{\"{u}}ter, and Hermann Ney.
\newblock Mean-normalized stochastic gradient for large-scale deep learning.
\newblock In {\em ICASSP}, 2014.

\bibitem{2016_NIPS_Wisdom}
Scott Wisdom, Thomas Powers, John Hershey, Jonathan Le~Roux, and Les Atlas.
\newblock Full-capacity unitary recurrent neural networks.
\newblock In {\em NIPS}, pages 4880--4888. 2016.

\bibitem{2017_Corr_Yu}
Adams~Wei Yu, Qihang Lin, Ruslan Salakhutdinov, and Jaime~G. Carbonell.
\newblock Normalized gradient with adaptive stepsize method for deep neural
  network training.
\newblock {\em CoRR}, abs/1707.04822, 2017.

\bibitem{2016_CoRR_Zagoruyko}
Sergey Zagoruyko and Nikos Komodakis.
\newblock Wide residual networks.
\newblock In {\em BMVC}, 2016.

\end{thebibliography}

\end{document}